\documentclass[10pt,twocolumn,letterpaper]{article}

\usepackage{iccv}
\usepackage{times}
\usepackage{epsfig}
\usepackage{graphicx}
\usepackage{amsmath}
\usepackage{amssymb}
\usepackage[font=small]{caption}
\usepackage{subcaption}
\usepackage{booktabs}
\usepackage{tikz}
\usepackage{pgfplots}
\usepackage{array}
\usepackage{xcolor}
\usepackage[color]{changebar}
\usepackage{microtype}
\usepackage[export]{adjustbox}

\iccvfinalcopy
\usepackage[pagebackref=false,breaklinks=true,colorlinks,letterpaper=true,bookmarks=false]{hyperref}

\newcolumntype{L}[1]{>{\raggedright\arraybackslash}p{#1}}

\hypersetup{%
  pdftitle={GazeDPM: Early Integration of Gaze Information in Deformable Part Models},
  pdfauthor={Iaroslav Shcherbatyi, Andreas Bulling, Mario Fritz},
  bookmarksnumbered,
  pdfstartview={FitH},
  colorlinks,
  citecolor=black,
  filecolor=black,
  linkcolor=black,
  urlcolor=black,
  breaklinks=true,
}

\cbcolor{red}

\def\iccvPaperID{548} %

\ificcvfinal\fi
\begin{document}

\title{GazeDPM: Early Integration of Gaze Information in Deformable Part Models}

\author{
Iaroslav Shcherbatyi$^{1,2}$
\and
Andreas Bulling$^1$\\
\and
Mario Fritz$^2$
\and
$^1$Perceptual User Interfaces Group, $^2$Scalable Learning and Perception Group\\
Max Planck Institute for Informatics, Saarbr\"ucken, Germany\\
{\tt\small \{iaroslav,bulling,mfritz\}@mpi-inf.mpg.de}
}

\newcommand{\abbrev}{GazeDPM}

\maketitle
\begin{abstract}

An increasing number of works explore collaborative human-computer systems in which human gaze is used to enhance computer vision systems.
For object detection these efforts were so far restricted to late integration approaches
that have inherent limitations, such as increased precision without increase in recall.
We propose an early integration approach
in a deformable part model, which constitutes a joint formulation over gaze and visual data.
We show that our \abbrev~method improves over the state-of-the-art DPM baseline by 4\% and a recent method for gaze-supported object detection by 3\% on the public POET dataset.
Our approach additionally provides introspection of the learnt models, can reveal salient image structures, and allows us to investigate the interplay between gaze attracting and repelling areas, the importance of view-specific models, as well as viewers' personal biases in gaze patterns.
We finally study important practical aspects of our approach, such as the impact of using saliency maps instead of real fixations, the impact of the number of fixations, as well as robustness to gaze estimation error.

\end{abstract}

\section{Introduction}

\begin{figure}[t]
    \centering
    \includegraphics[width=0.95\columnwidth]{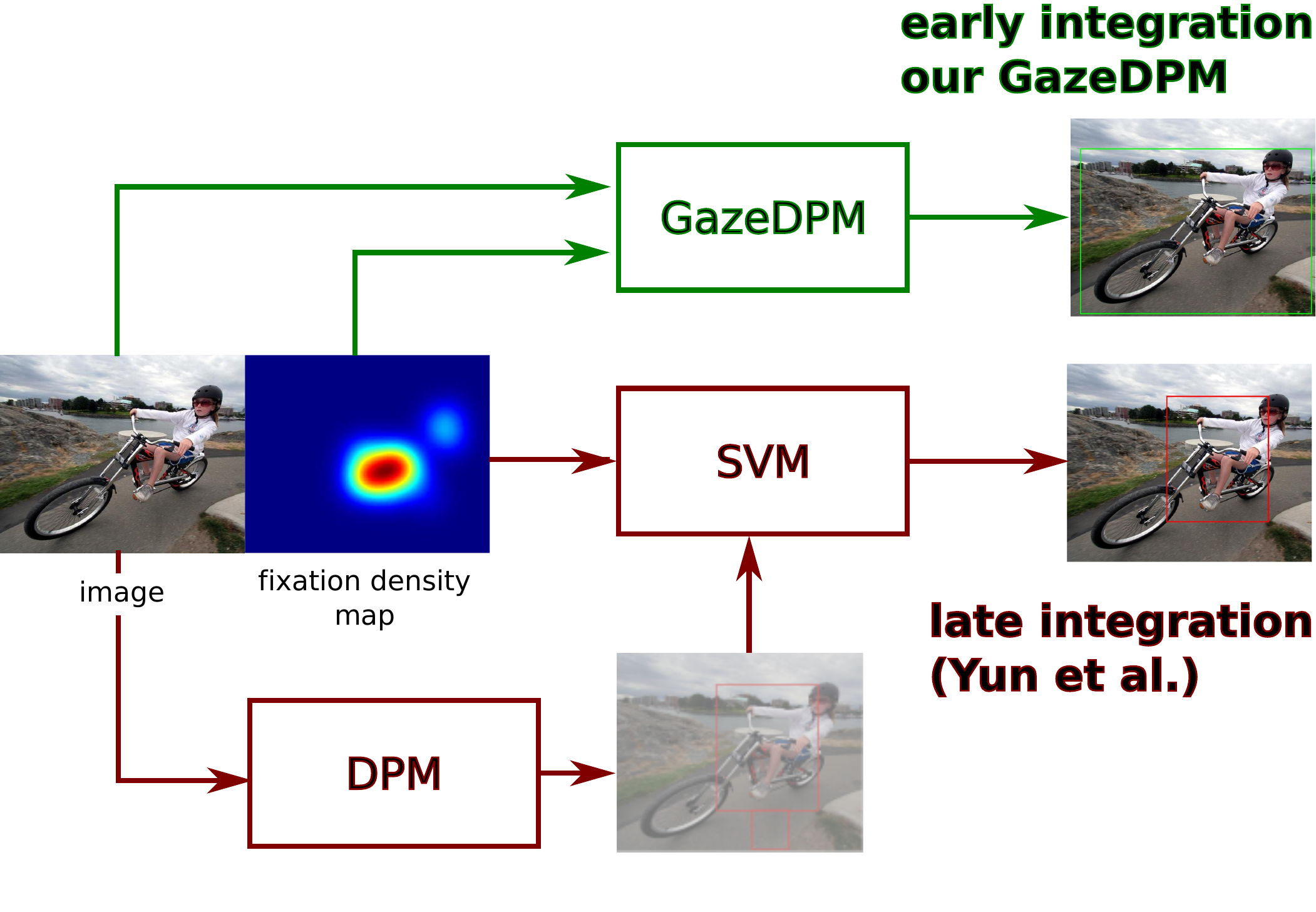}
    \caption{A recent late integration approach for gaze-supported object detection~\cite{yun2013studying} learns from image and gaze information separately (bottom). In contrast, our \abbrev~method enables early integration of gaze and image information (top).}
    \label{fig:teaser}
\end{figure}

Across many studies, human gaze patterns were shown to reflect processes of cognition, such as intents, tasks, or cognitive load, and therefore represent a rich source of information about the observer.
Consequently, they have been successfully used as a feature for predicting the user's internal state, such as user context, activities, or visual attention~\cite{bulling11_pcm,bulling11_pami,bulling13_chi,majaranta14_apc}.
Recent advances in eye tracking technology~\cite{tsukada2011illumination,Kassner14_ubicomp,xucong15cvpr,wood14_etra,Zhang14_AVIb} open up a wide range of new opportunities to advance human-machine collaboration (e.g. \cite{sattar15_cvpr}) or aid computer vision tasks, such as object recognition and detection~\cite{yun2013studying,papadopoulos2014training,karthikeyan2013and}.
The overarching theme is to establish collaborative human-machine vision systems in which part of the processing is carried out by a computer and another part is performed by a human and conveyed to the computer via gaze patterns, typically in the form of fixations.

Yun et al.\ recently applied this approach to object detection and showed how to improve performance by re-scoring detections based on gaze information~\cite{yun2013studying}.
However, image features and gaze information were processed independently and only the outputs of the two pipelines were fused.
This constitutes a form of late integration of both modalities and comes with inherent limitations.
For example, the re-scoring scheme can improve precision but cannot improve recall.
Also, exploitation of dependencies between modalities is limited as two separate models have to be learned.

In contrast, we propose an early integration scheme using a joint formulation over gaze and visual information (see~\autoref{fig:teaser}).
We extend the deformable part model~\cite{felzenszwalb2010object} to combine deformable layouts of gradient and gaze patterns into a \abbrev~model.
This particular model choice allows for rich introspection into the learned model and direct comparison to previous work employing late integration.
Our analysis reveal salient structures, interplay between gaze attracting and repelling areas, importance of view-specific models as well as personal biases of viewers. 
As we have highlighted the emerging opportunities of applying such collaborative schemes in applications, we further study and quantify important practical aspects, such as benefits of human gaze data over saliency maps generated from image data only, temporal effects of such a collaborative scheme, and noise in the gaze measurements.

The specific contributions of this work are threefold:
First, we present the first method for early integration of gaze information for object detection based on a deformable part models formulation.
In contrast to previous late integration approaches where gaze information is only used to re-score detections, we propose a joint formulation over gaze and visual information.
Second, we compare our method with a recent late integration approach~\cite{yun2013studying} on the publicly available POET dataset~\cite{papadopoulos2014training}.
We show that our early integration approach outperforms the late integration approach in terms of mAP by 3\% and provides a deeper insight into the model and properties of the data. 
Third, we present and discuss additional experiments exploring important practical aspects of such a collaborative human-computer systems using gaze information for improved object detection.

\section{Related Work}

Our method is related to previous works on 1) deformable part models for object detection, 2) visual saliency map estimation, and 3) the use of gaze information in collaborative human-computer vision systems.

\paragraph{Deformable Part Models}

One of the most successful approaches for object detection over the last decade is the deformable part models~\cite{felzenszwalb2010object}.
Deformable part models constitute of a set of linear filters that are used to detect coarse representation of an object and refine detections using filters that respond to specific details of objects being detected.
Because of their simplicity and ability to capture complex object representations, many extensions of deformable parts models have been proposed in literature~\cite{pascal-voc-2012}, where usage of more complex pipelines or better features allows to improve detection performance.
Recently, improved detection performance was demonstrated using neural network based approaches~\cite{ILSVRCarxiv14,girshick14CVPR}.
In this work we opted to build on deformable part models because they allow for better introspection and have the potential to better guide future work on the exploration of gaze information in computer vision.
Introspection in deep architecture is arguably more difficult and topic of ongoing research~\cite{simonyan14deep,zeiler14eccv}.

\paragraph{Visual Saliency Map Estimation}

Saliency estimation and salient object detection algorithms can be used for scene analysis and have many practical applications \cite{itti1998model}. For example, they allow to estimate probability of observer fixating on some area in image and thus allow to model which parts of depicted scene attracts attention of a human observer.
A variety of different saliency approaches were developed, like 
graph based visual saliency \cite{harel2006graph},
boolean map approach \cite{zhang2013saliency},
and recent approaches using neural networks \cite{vig2014large} 
that allow to estimate saliency maps or detect most salient objects for a given scene or video segment. 

To evaluate saliency algorithms, many datasets containing images and eye tracking information from a number of observers are available. For example, eye tracking data is available in \cite{judd2009learning, yun2013studying} for free viewing task, in a large POET dataset \cite{papadopoulos2014training} for a visual search task, and in \cite{li2009dataset} for evaluation of saliency algorithms on video sequences.
The acquisition of gaze data can be achieved via a wide range of methods \cite{tsukada2011illumination,Kassner14_ubicomp,xucong15cvpr,wood14_etra,Zhang14_AVIb} -- which is not part of our investigation, although we do evaluate robustness withi respect to noise in the gaze data.
We evaluate our work on the existing POET dataset and investigate in how far saliency maps can substitute real gaze data in our approach.

\paragraph{Collaborative Human-Computer Vision Systems}

There has recently been an increasing interest in using gaze information to aid computer vision tasks.
For example, fixation information was used to perform weakly supervised training of object detectors~\cite{papadopoulos2014training,karthikeyan2013and}, analysing pose estimation tasks~\cite{mps13iccv}, inferring scene semantics~\cite{subramanian2011can}, detecting actions~\cite{mathe2014multiple}, or predicting search tasks~\cite{sattar15_cvpr}.
Our approach more specifically targets the use of gaze data for object detection.
The most closely related work to ours is~\cite{yun2013studying}, where gaze information was used to re-score detections produced by a deformable part model.
In contrast, the proposed \abbrev~approach integrates gaze information directly into deformable part models and therefore provides a joint formulation over visual and gaze information. 
We further consider saliency maps as a substitute for real gaze data.
Similar ideas can be found in~\cite{rutishauser2004bottom, moosmann2006learning}, where it was demonstrated that saliency maps can be used to improve object detection performance.

\section{Gaze-Enabled Deformable Part Models}
\begin{figure*}[t]
\centering
\includegraphics[width=0.8\linewidth]{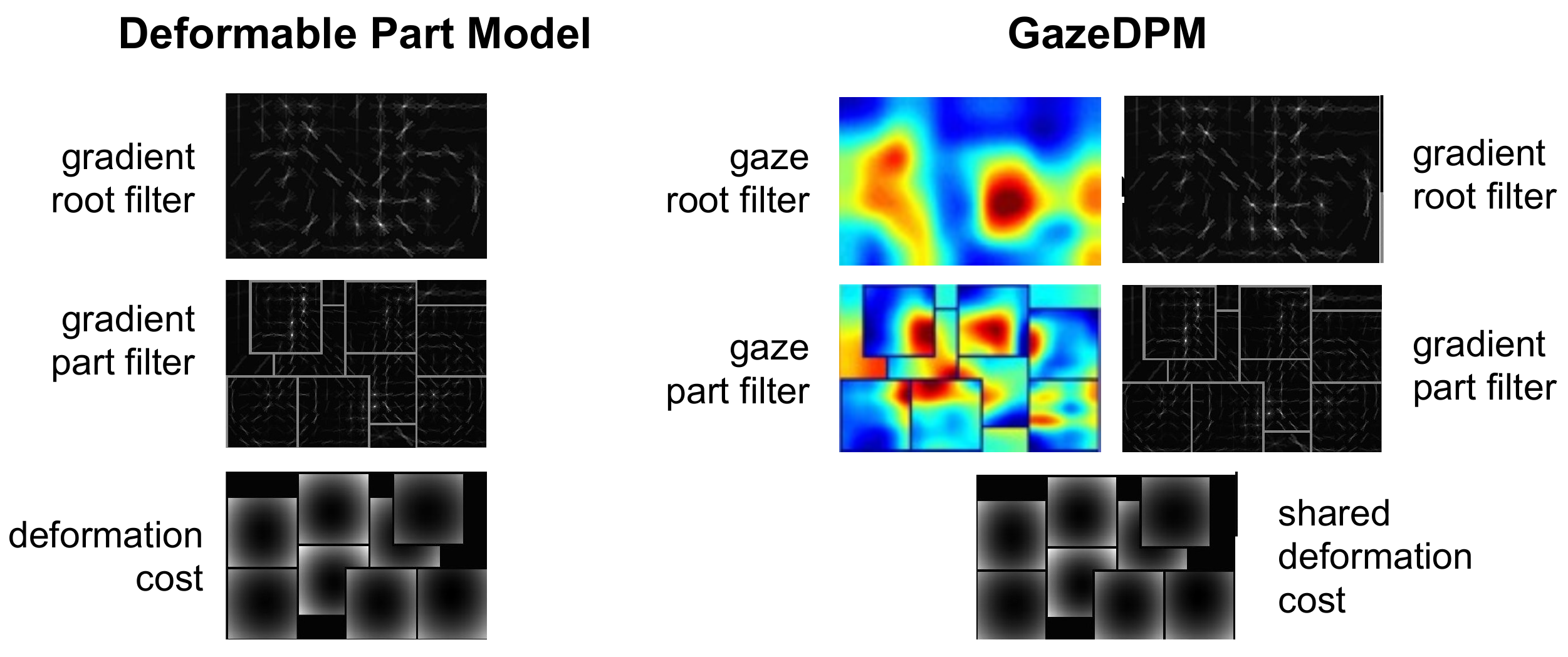}       
\caption{Comparison of vision-only deformable part model (DPM) on the left and gaze-enabled deformable part model (GazeDPM) on the right.} 
\label{fig-POET-concept-figure}
\end{figure*} 

To formulate a joint model of visual and gaze information we build on the established deformable part models (DPM) \cite{felzenszwalb2010object}. In contrast to recent developments in deep learning, this particular model allows for better model introspection.
Deformable part models predict bounding boxes for object categories in an image (e.g.\ a bicycle) based on visual features. In this section we describe necessary background and our extension of deformable part models towards our new \abbrev~formulation, which is used in further sections for gaze-enabled object detection. 

\subsection{Visual Feature Representation}

DPMs use feature maps for object detection in images.
Feature maps are arrays of feature vectors in which every feature vector contains local information that corresponds to some patch in an image (e.g.\ the average direction or magnitude of derivative).
To enable DPMs to detect objects on different scales, a feature pyramid is used, which consists of feature maps computed from an image on different scales.
Throughout this work we use a 31 dimensional feature representation as described in~\cite{felzenszwalb2010object}  that we obtained by analytical reduction of HOG features~\cite{dalal2005histograms} in addition to gaze features.

\subsection{Fixation Density Maps}

\begin{figure}[t]
\centering
\includegraphics[width=\columnwidth]{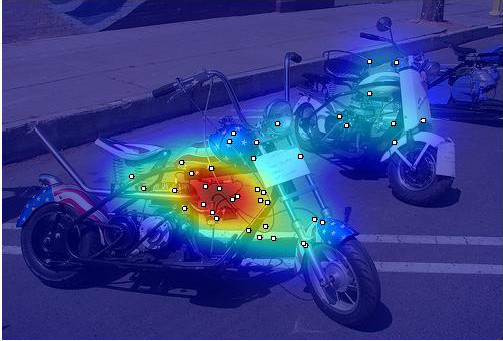}
\caption{Example fixation density map overlayed on the corresponding image from the POET dataset. White dots represent fixations by different observers; color close to red indicates high fixation density, and close to blue small density. Note that gaussians around each fixation are weighted by fixation duration.}
\label{fig-POET-fixation-density-map-example}
\end{figure}

Gaze information is available as two-dimensional coordinates of observers' fixations on an image as well as their duration.
We encode sequences of these fixations into fixation density maps that proved useful for many tasks.
This representation was used in prior work~\cite{yun2013studying} and we follow the same approach here.
For every image a fixation density map is obtained by pixelwise summation of values of weighted Gaussian functions placed at every fixation position in an image and normalizing the resulting map to values in a range from 0 to 1. Every Gaussian function in a sum corresponds to normal distribution function, with mean equal to fixation coordinates and covariance matrix as a diagonal matrix with values of $ \sigma^2 $ on diagonal, where $ \sigma $ is selected to be 7~\% of image height. Weight of the Gaussian function is selected to be corresponding fixation duration. Normalization of fixation map is obtained by dividing the values of sum of weighted Gaussian functions by its maximum value. 
This representation is equally applicable if real gaze fixations are not available and a saliency map algorithm is used as a substitute~\cite{itti1998model, harel2006graph, zhang2013saliency}.
These methods also produce a density map that tries to mimic an actual fixation density map produced by real fixation data.
A sample fixation density map obtained from real fixations is shown in \autoref{fig-POET-fixation-density-map-example}, where it is overlayed onto the corresponding image.

\subsection{Deformable Part Models}

Deformable parts models are star models defined by a root filter that is used to detect the coarse, holistic representation of the whole object, and part filters that are used to detect individual parts of an object.
Root filter detections are used to determine an anchor position and the score of root and part filters together with deformation coefficients are used to compute a detection score with latent part placement.
The score maps of the root and part filters are computed by convolution of feature maps in the feature pyramid.

\begin{figure*}[t]
\centering
\begin{minipage}[b]{.5\linewidth}
  \centering
  \begin{tabular}{ccc}
    \includegraphics[  scale=0.1]{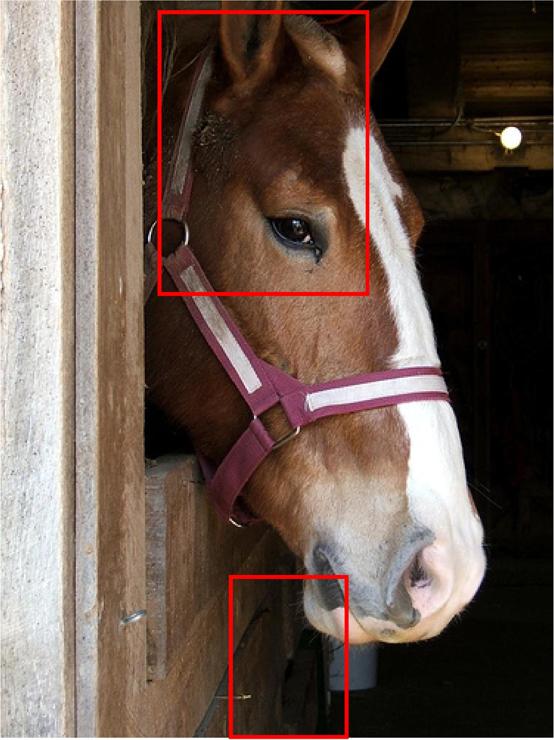}&
    \includegraphics[  scale=0.1]{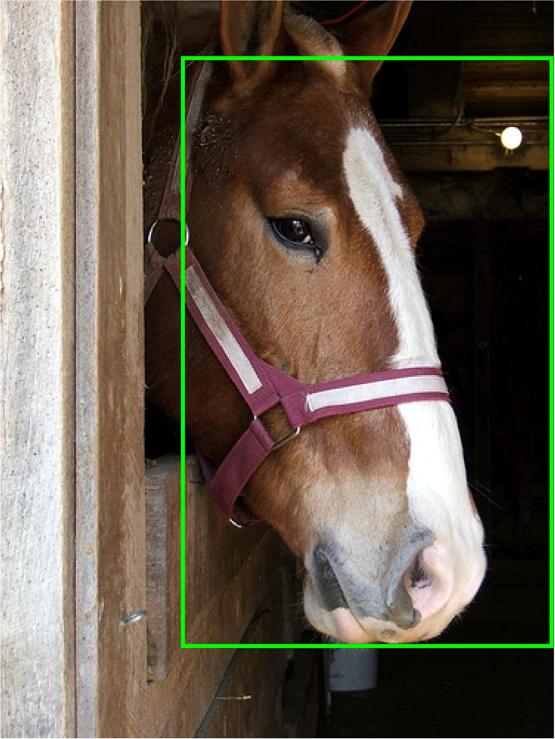}&
    \includegraphics[  scale=0.1]{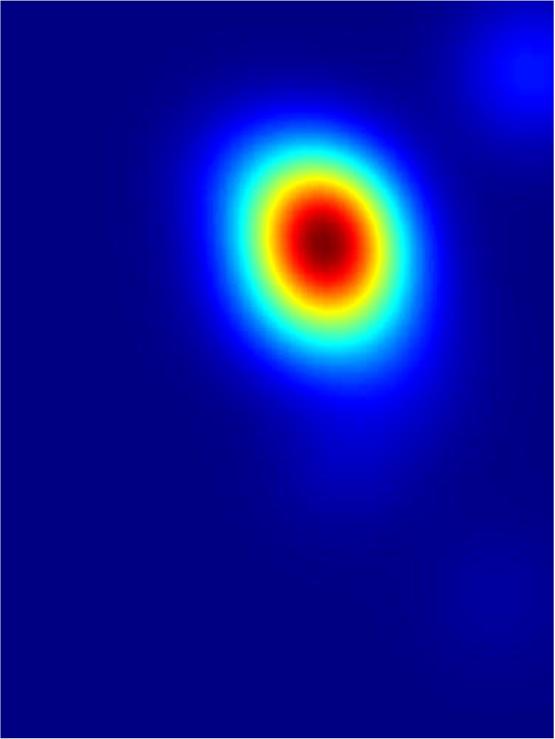}\\
  \end{tabular}
  \subcaption{``horse''}\label{subfig:boat_comparison_2}
\end{minipage}%
\begin{minipage}[b]{.5\linewidth}
  \centering
  \begin{tabular}{ccc}
    \includegraphics[  scale=0.09]{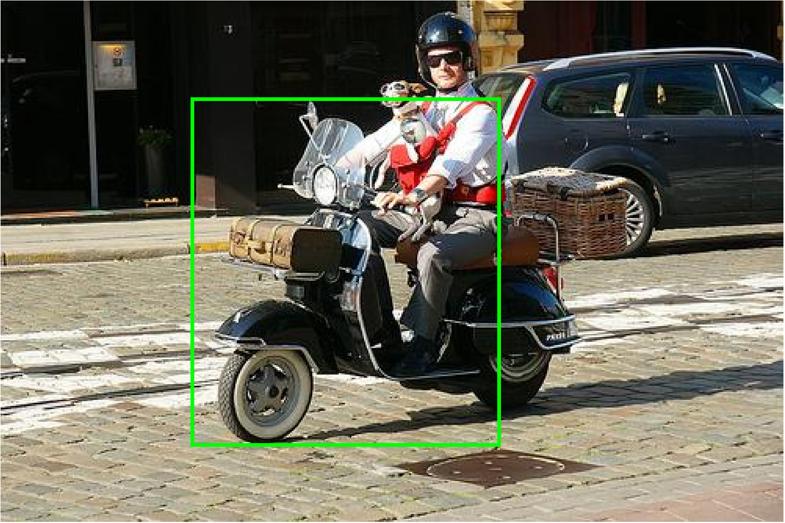}&
    \includegraphics[  scale=0.09]{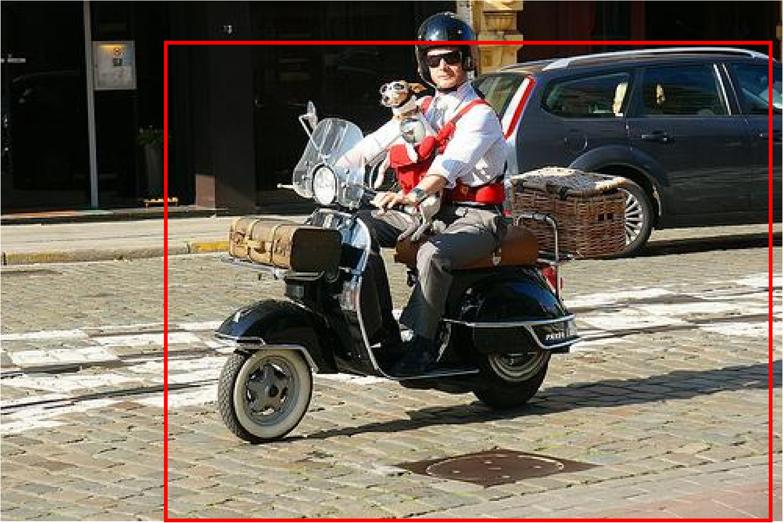}&
    \includegraphics[  scale=0.09]{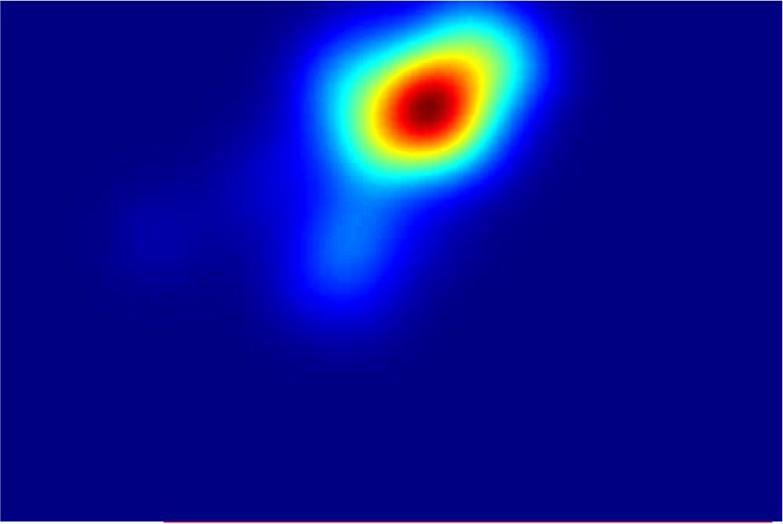}\\
  \end{tabular}
  \subcaption{``motorbike''}\label{subfig:motorbike_comparison_5}
\end{minipage}
\begin{minipage}[b]{.5\linewidth}
  \centering
  \begin{tabular}{ccc}
    \includegraphics[ scale=0.1]{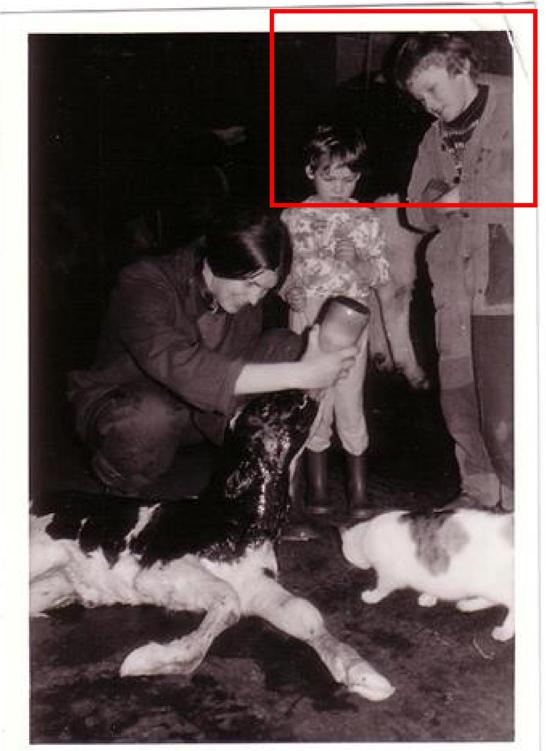}&
    \includegraphics[scale=0.1]{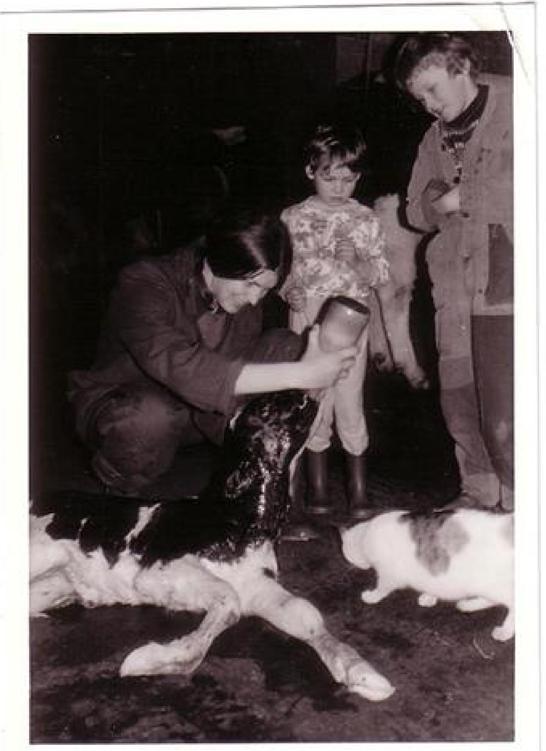}&
    \includegraphics[scale=0.1]{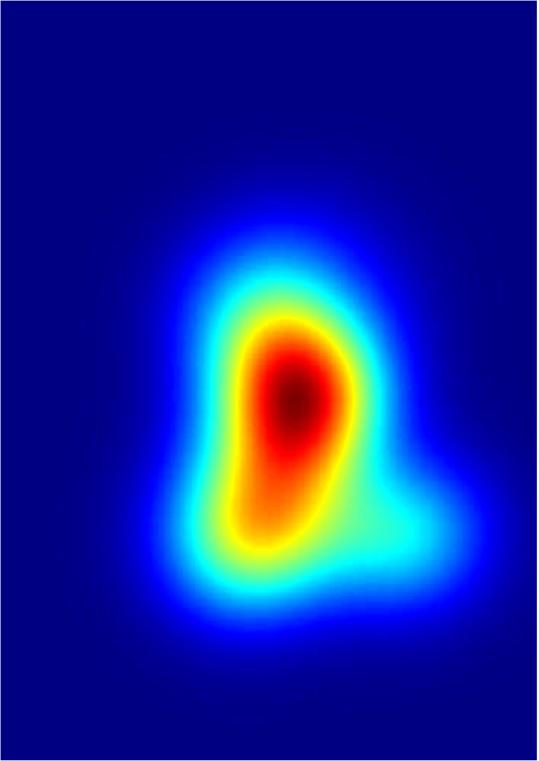}\\
  \end{tabular}
  \subcaption{``cow''}\label{subfig:cow_comparison_1}
\end{minipage}%
\begin{minipage}[b]{.5\linewidth}
  \centering
  \begin{tabular}{ccc}
    \includegraphics[scale=0.1]{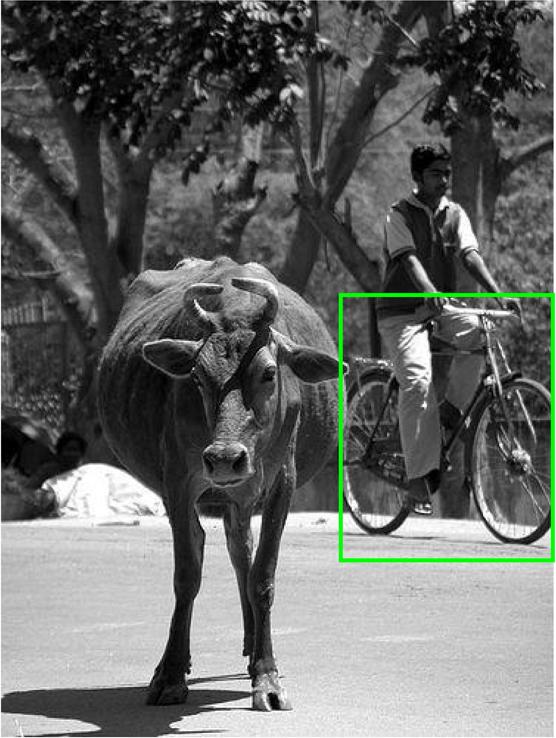}&
    \includegraphics[scale=0.1]{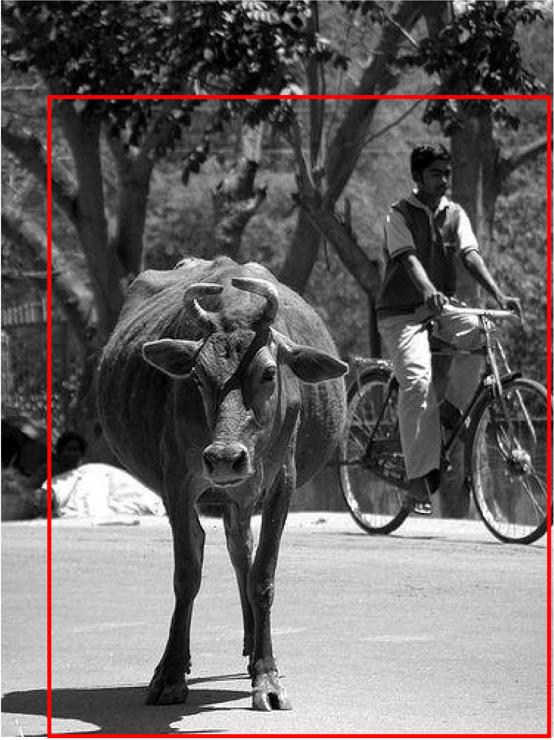}&
    \includegraphics[scale=0.1]{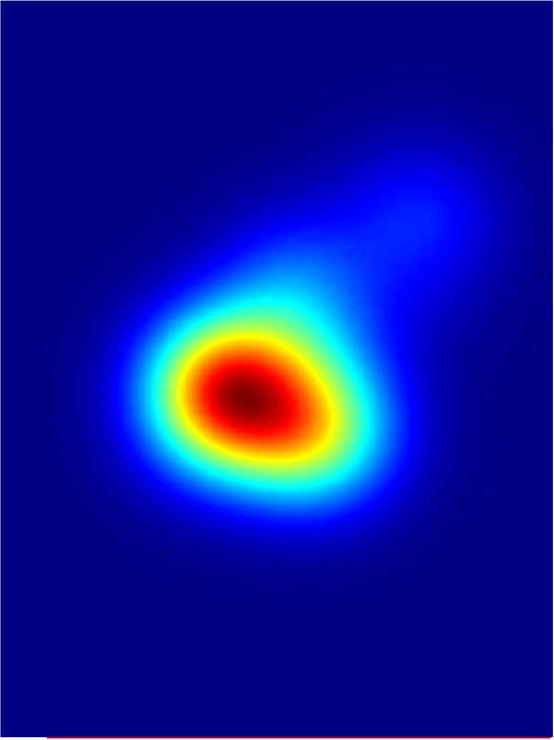}\\
  \end{tabular}
  \subcaption{``bicycle''}\label{subfig:bicycle_comparison_5}
\end{minipage}
\caption{Example detections on images of four different classes from the POET dataset~\cite{papadopoulos2014training}. For every triple of images, left is the image with detections of the original DPM, center is the image with \abbrev~detections, and right is the density map generated from the fixations. True positive detections are shown in green, false positive detection in red. }
\label{fig:cats-figure} 
\end{figure*}

Each part filter is anchored at some position relative to the root filter.
Let $ P $ denote set of all possible locations in image.
For a given deformable part model and some location $ p_0 \in P $ in an image, the overall score of detection in this location is given by
\begin{align}
s(p_0) = r_0(p_0) + \max\limits_{p_1 ... p_n} \sum_{i \in [n]} r_i(p_i) - d_i(p_0, p_i) ,
\end{align}
where  $ p_i \in P, i \in [n], n \in \mathbb{N} $, function $ s(p_0) $ is a score of DPM positioned in image at position $ p_0 $, $ r_0 $ is a response of root filter at $ p_0 $, $ r_i(p_i) $ is a response of filter that corresponds to the part that is located at position $ p_i $, and $ d_i(p_0, p_i) $ is a displacement penalty.

A sliding window approach is used to detect objects with this model.
For every position in the image and at every scale, an optimal placement of parts is determined by maximizing the score function.
If the found score is above a threshold, the hypothesis that object is present in bounding box is accepted.
As part placements are latent, a latent SVM formulation is used for training~\cite{felzenszwalb2010object}. DPMs can then be expressed as a classifier of the following form:
\begin{align}
f_\beta(x) = \max\limits_{z \in Z(x)} \langle \beta, \Phi(x,z) \rangle ,
\label{eq:dpm-classifier}
\end{align}
where $ x $ is an image feature map, $ Z(x) $ is a set of all possible sliding windows in an image, $ \beta $ is a vector that contains weights of all linear filters weights and displacement costs, and $ \Phi(x,z) $ is a features subset that corresponds to some sliding window $ z \in Z(x) $. Then, training DPM corresponds to finding parameters $\beta$ of linear filters in DPM such that they minimize objective
\begin{align}
L(\beta)=||\beta||_2^2 + C \sum_{i \in m} \max(0,1-y_i f_\beta(x_i)) ,
\end{align}
where $ y_i $ is a label that indicates if an instance of a class is present in image $x_i$ and there are $ m \in \mathbb{N} $ such images available. For more information regarding optimization of above function as well as on other details please refer to \cite{felzenszwalb2010object}.

\subsection{Integration of Gaze Information}

We extend the original DPM formulation by adding additional parts that are trained on a new fixation density map feature channel:
\begin{align}
f_\beta(x) = \max\limits_{z \in Z(x)} \langle \beta, \Phi(x,z) \rangle + \langle \beta', \Phi'(x,z) \rangle  ,
\label{eq:gazedpm-classifier}
\end{align}
where $ \beta' $ corresponds to parameters of linear filters that are applied to fixation features $ \Phi'(x,z) $.
We call the resulting extended DPM ``\abbrev''.
We refer to this method of integration of gaze information as ``early integration'' given that fixation data is directly used in the DPM.
This is in contrast to a recent work~\cite{yun2013studying} that used a ``late integration approach'' by using fixation information not directly in the DPM but to refine its detections.
\autoref{fig-POET-concept-figure} provides visualisations comparing these two different integration methods.
The overall detection score for \abbrev~model applied at some location $ p_0 $ in an image can be computed as 

\begin{align}
s'(p_0) &= R_0(p_0) + \max\limits_{p_1 ... p_n} \sum_{i \in [n]} R_i(p_i) - d_i(p_0, p_i),
\end{align}
where $ R_i(p_i) $ 
\begin{align}
R_i(p_i) &= r_i(p_i) + r_i'(p_i), \\
i &\in \{0, ... ,n\},
\end{align}
denotes joint response of linear filter $ r_i'(p_i) $ applied to gaze features and response of linear filter $ r_i(p_i) $ applied to image features at position $ p_i \in P $ and $ i = 0 $ denotes root filter, $ i \in \{ 1, ..., n\} $ denotes part filter.

\subsection{Implementation}

We implemented our \abbrev~model based on the MATLAB implementation of the original deformable part models provided with~\cite{felzenszwalb2010object}. In order to ensure reproducability and stimulate research in this area we will make code and models publicly available at time of publication.
In the following, we provide experimental evaluation of our \abbrev~model in different settings, compare to prior work, and provide additional insights and analysis into the learnt models.

\section{Experiments on POET}

We first compare our \abbrev~method to the early integration approach proposed in~\cite{yun2013studying}. 
All of the experiments presented in this section were performed on the Pascal Objects Eye Tracking (POET) dataset~\cite{papadopoulos2014training}.
This dataset contains eye tracking data for 10 classes of the original Pascal VOC 2012 dataset.
Eye tracking data was collected from observers whose task was to find one of Pascal classes present in the image (visual search task).
We split the dataset into training and testing sets of approximately equal number of class instances (approx. 3000 testing and training images) and use this split throughout all of experiments below, unless specified otherwise.

For evaluating the performance of the models we use the evaluation code provided with the VOC dataset \cite{pascal-voc-2012}.
For all experiments we used two aspect ratio clusters for DPM detectors and default thresholds.
We found experimentally that two clusters yielded the best performance for stock DPMs and our modification on POET dataset and therefore used these settings throughout the experiments. 
For more detailed account of experimental results described in this section please refer to supplementary material.

\subsection{Early vs Late Integration}

\def\arraystretch{1.3}
\begin{table*}[t]
\begin{center}
\begin{tabular}{lllL{1.5cm}lllL{1.5cm}lllll}
   & \textbf{DPM}  & \textbf{\cite{yun2013studying}} & \textbf{\abbrev} & \multicolumn{4}{c}{\textbf{Amount of noise}} & \multicolumn{5}{c}{\textbf{Participant-specific fixations}} \\ \hline
\textbf{Class} & original & late & early  & $0.5\sigma^2$ & $\sigma^2$    & $1.5\sigma^2$ & $2\sigma^2$   & P1        & P2        & P3        & P4        & P5 \\ \toprule
cat              & 23.9          & 24.0                            & 40.2             & 39.0          & 33.1          & 33.0          & 29.1          & 36.3          & 34.4          & 35.4          & 36.1          & 33.4 \\ \hline   
cow              & 22.6          & 22.6                            & 24.9             & 20.3          & 21.1          & 18.6          & 21.9          & 21.3          & 22.7          & 21.2          & 19.1          & 20.1 \\ \hline   
dog              & 14.7          & 15.2                            & 28.2             & 23.5          & 23.2          & 15.5          & 15.8          & 24.6          & 18.5          & 22.8          & 22.5          & 23.4 \\ \hline   
horse            & 43.9          & 44.0                            & 46.0             & 44.5          & 42.6          & 40.7          & 40.7          & 43.0          & 46.5          & 41.8          & 43.9          & 43.8 \\ \hline   
aeroplane        & 41.8          & 42.3                            & 40.6             & 42.4          & 42.2          & 44.2          & 44.8          & 40.4          & 38.9          & 42.4          & 39.9          & 43.8 \\ \hline   
bicycle          & 53.5          & 53.8                            & 53.5             & 53.9          & 51.9          & 52.6          & 52.8          & 53.4          & 52.6          & 52.6          & 53.0          & 52.5 \\ \hline   
boat             & 8.4           & 8.4                             & 9.3              & 10.1          & 8.7           & 7.2           & 8.8           & 10.0          & 10.3          & 7.1           & 9.3           & 7.8 \\ \hline    
diningtable      & 19.8          & 21.8                            & 30.0             & 30.8          & 15.0          & 13.0          & 23.1          & 24.3          & 26.2          & 18.5          & 27.9          & 26.1 \\ \hline   
motorbike        & 48.5          & 48.7                            & 45.9             & 46.1          & 46.4          & 47.1          & 46.4          & 44.3          & 43.4          & 44.2          & 44.4          & 46.1 \\ \hline   
sofa             & 26.7          & 27.4                            & 28.5             & 32.8          & 31.1          & 25.8          & 24.3          & 29.1          & 27.1          & 29.2          & 29.4          & 23.7 \\ \bottomrule   
\textbf{Average} & \textbf{30.4} & \textbf{30.8}                   & \textbf{34.7}    & \textbf{34.3} & \textbf{31.5} & \textbf{29.8} & \textbf{30.8} & \textbf{32.7} & \textbf{32.1} & \textbf{31.5} & \textbf{32.5} & \textbf{32.1}    
\end{tabular}
\end{center}
\caption{Performance comparison of all three methods (original DPM~\cite{felzenszwalb2010object}, late integration~\cite{yun2013studying}, and our \abbrev) on the POET dataset. For all modifications two aspect ratio clusters were used. P1 ... P5 means only fixations from that specific participant were used to generate fixation density maps. Columns with multiples of $ \sigma^2 $ denote performance with fixation density maps generated with different amounts of noise to simulate the influence of low-accuracy gaze estimation settings.}
\label{table-POET-results}
\end{table*}

We re-implemented the late integration method proposed in~\cite{yun2013studying} with assistance of the authors.
Evaluation results are provided in~\autoref{table-POET-results}.
For comparison, we also show the unmodified DPM performance.
As can be seen from the table, our reimplementation of the late integration scheme achieves an improvement of $0.4\%$ which is consistent with the results published in~\cite{yun2013studying}.
Our \abbrev~achieves an overall performance of $34.7\%$, which is a $3.9\%$ improvement over the late integration \cite{yun2013studying} and $4.3\%$ improvement over the DPM baseline.
These results provide evidence for the benefits of an early integration scheme and joint modelling of visual and fixation features.

Notice that for the late integration of gaze information \cite{yun2013studying} one needs to have a baseline dpm model, which should be trained on images other than those that are used as training and testing set for gaze classifier procedure. In the experiment described in this section we however used training set for both gaze classifier procedure and dpm training. To account for this, we established a more rigorous comparison setup which is more favorable for late integration, which for brewity we do not describe here (see supplement), but for which we still get an improvement with GazeDPM of 3 \% mAP compared to late integration.

\subsection{Further Analysis of \abbrev}

To gain deeper insights, we further analyzed and visualized the \abbrev~models that we trained.

\subsubsection{Analysis of Example Detections}

Example detections and error cases are shown in~\autoref{fig:cats-figure}. 
Given that fixation information is quite informative regarding the class instance, 
it allows our method to obtain more true positive detections (see~\autoref{subfig:boat_comparison_2}) and remove some false positives (see~\autoref{subfig:cow_comparison_1}).
We observed that there was a tendency towards bounding boxes covering most of fixations as can be seen in~\autoref{subfig:motorbike_comparison_5}.
This indicates that the model has learnt that fixations are a strong indicator for object presence, which is exploited by our \abbrev~method. 
This assumption can be violated, as observers also produce spurious fixations in search tasks.
For some cases such fixation related to exploration of the image can create false positives as can be seen in~\autoref{subfig:bicycle_comparison_5}.
\subsubsection{Learning Salient Structures}

\begin{figure}[t]
\centering
\begin{minipage}[b]{0.2\textwidth}
  \centering
  \includegraphics[height = 2.5cm]{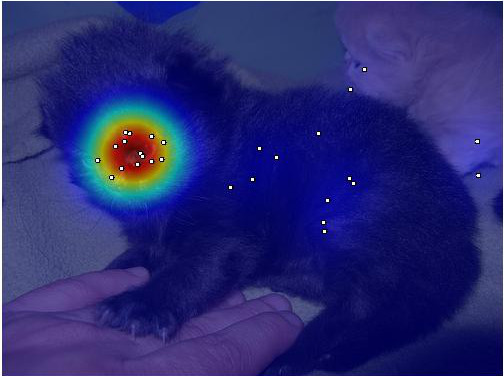}%
  \subcaption{}
  \label{subfig:struct-cat1}
\end{minipage}%
\centering
\begin{minipage}[b]{0.24\textwidth}
  \centering
  \includegraphics[height = 2.5cm]{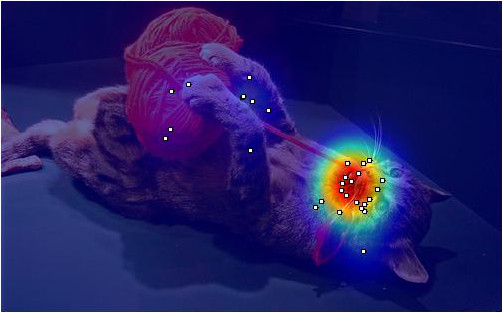}%
  \subcaption{}
  \label{subfig:struct-cat2}
\end{minipage}%
\caption{Example salient structures in images (a, b). Notice that people tend to not fixate on animal neck. Compare to learned weights of gaze filters in \autoref{fig:cat-distribution-components}}
\label{fig:struct-cat}
\end{figure}

When looking at some of the part visualizations and corresponding fixation density maps we realized that \abbrev~models are able to learn salient structures for different categories and aspect ratio clusters.
For example, for the cat category,~\autoref{fig:struct-cat} shows some example images with positions of observer fixations.
It is well known that people tend to fixate on heads of animals~\cite{yun2013studying} and this can be seen by visualizing the distribution of fixations for the cat class (see~\autoref{fig:cat-distribution-components}). We also found that people usually tend to not fixate on animal neck (see~\autoref{fig:struct-cat}).
This is reflected in the resulting \abbrev~models by a strong positive weight (see~\autoref{fig:cat-distribution-components}) which acts like a ``gaze attractor'' at location where the animal head is located, and by a strong negative weight which acts like a ``gaze repellent'' in the area where animal neck is located.
We like to draw the attention to the root filter of one component model  in the gaze density map with negative weight close to the neck of animal. In this way our \abbrev~model tries to exploit such salient structures, present in training data. Similar effects can be seen on gaze parts filters; However, as parts filters can be shifted, they appear to be located in such a way so as to account for different locations of peaks (animal head) in fixation map specific for an image.
Looking across the learnt filters, we see an interesting interaction between areas on the object that attract fixations and close-by regions that are not fixated.
The latter areas can be seen as ``gaze repellents'', which might be due a shadowing effect of the close-by attractor.
\subsubsection{Learning View-Specific Information}

\begin{table*}
\centering
\setlength{\tabcolsep}{2pt}
\begin{tabular}{cccccccc}
\textbf{\# of comp} & \textbf{Fixation density map} & \textbf{Gradient root} & \textbf{Gradient parts} &  \textbf{Gaze root} & \textbf{Gaze parts} & \textbf{Deformations}\\\toprule
1&
\begin{tabular}{cc}
\includegraphics[height = 0.08\linewidth]{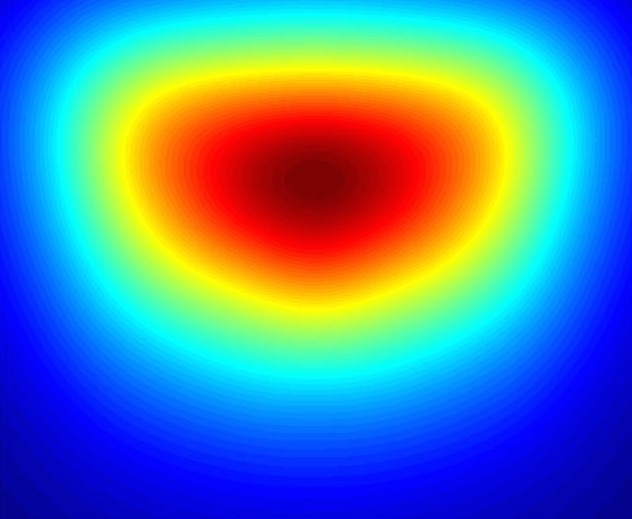}
\end{tabular}&
\begin{tabular}{cc}
\includegraphics[height = 0.08\linewidth]{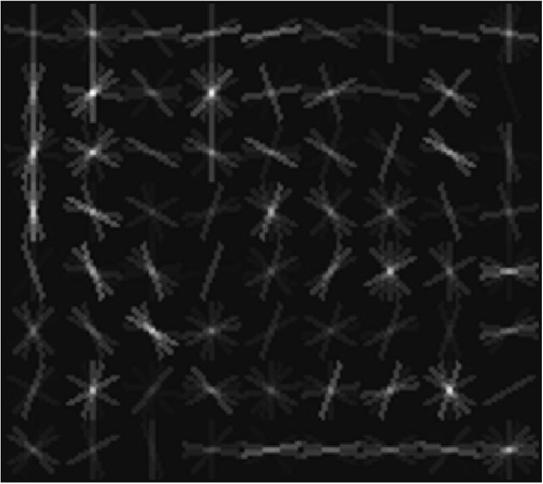} 
\end{tabular}&
\begin{tabular}{cc}
\includegraphics[height = 0.08\linewidth]{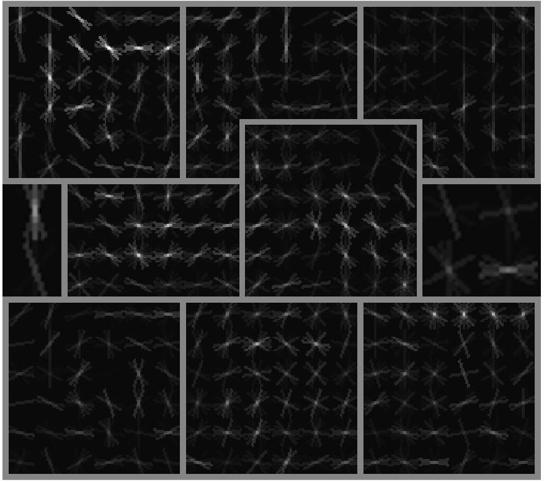}
\end{tabular}& 
\begin{tabular}{cc}
\includegraphics[height = 0.08\linewidth]{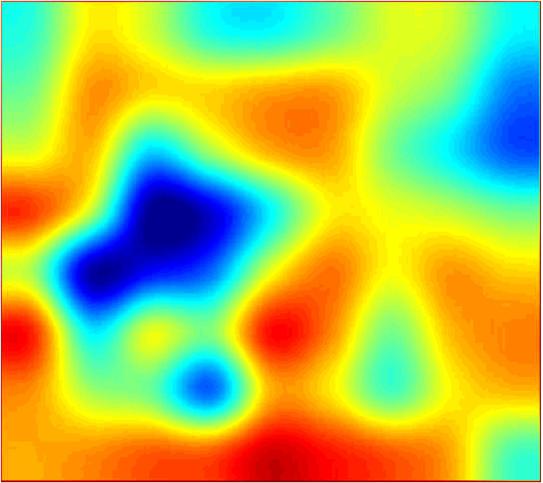}
\end{tabular}& 
\begin{tabular}{cc}
\includegraphics[height = 0.08\linewidth]{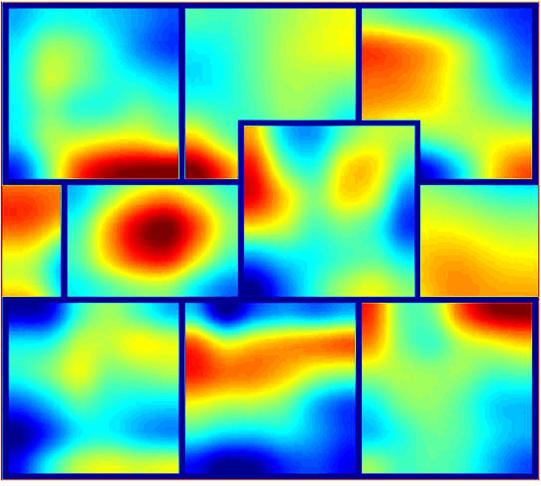}
\end{tabular}& 
\begin{tabular}{cc}
\includegraphics[height = 0.08\linewidth]{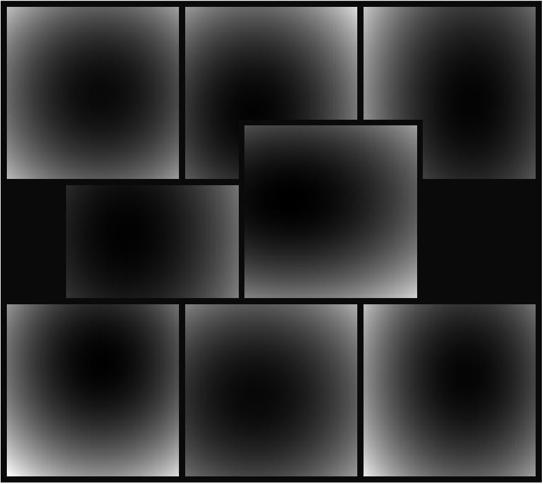}
\end{tabular}\\\midrule
2&
  \begin{tabular}{cc}
    \includegraphics[height = 0.08\linewidth]{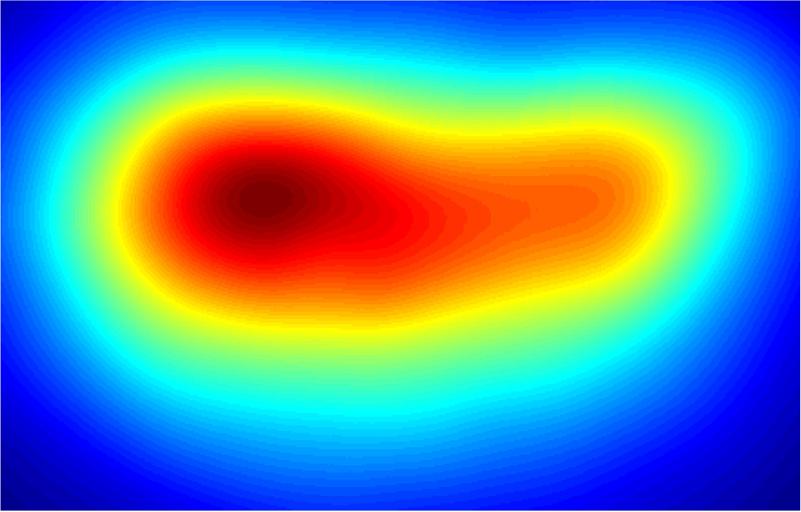}\\
    \includegraphics[height = 0.1\linewidth]{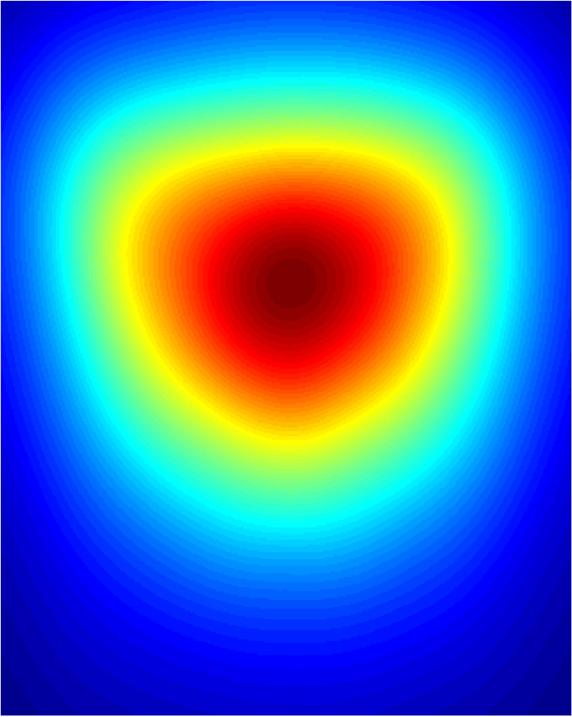}
\end{tabular}&
  \begin{tabular}{cc}
    \includegraphics[height = 0.08\linewidth]{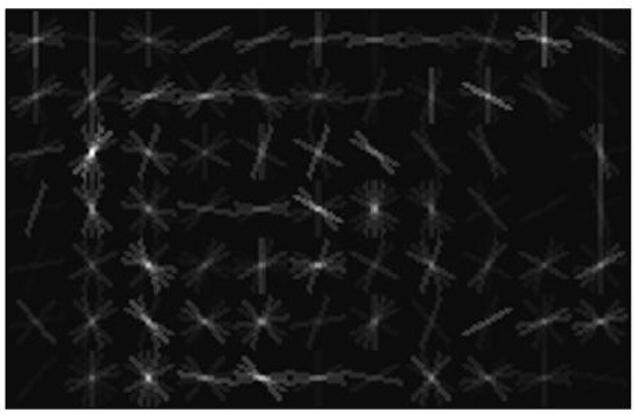}\\
    \includegraphics[height = 0.1\linewidth]{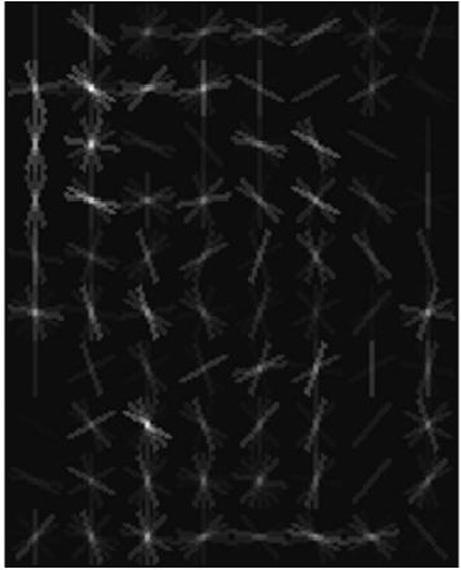}
\end{tabular}&
  \begin{tabular}{cc}
    \includegraphics[height = 0.08\linewidth]{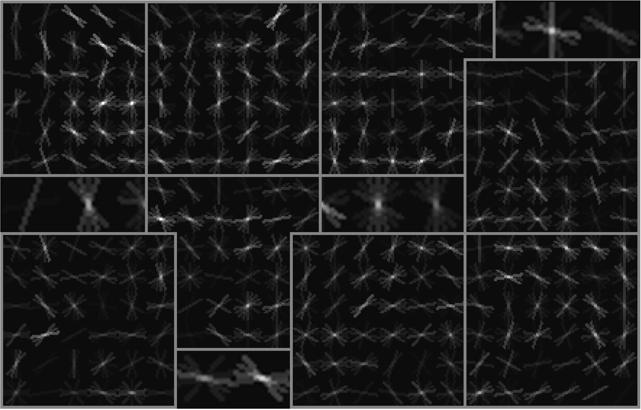}\\
    \includegraphics[height = 0.1\linewidth]{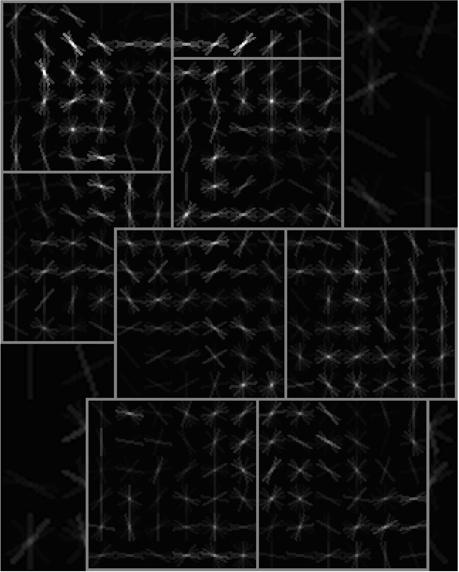}
\end{tabular}&
  \begin{tabular}{cc}
    \includegraphics[height = 0.08\linewidth]{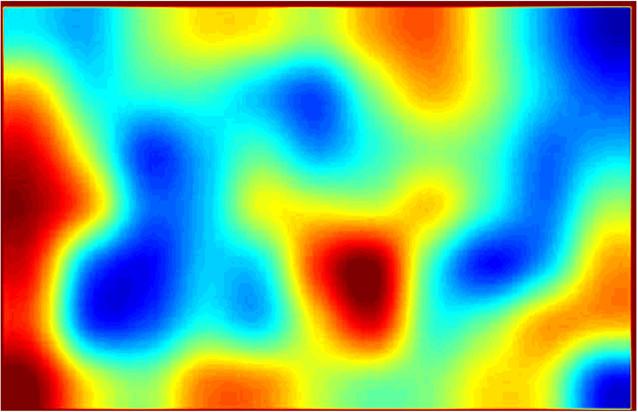}\\
    \includegraphics[height = 0.1\linewidth]{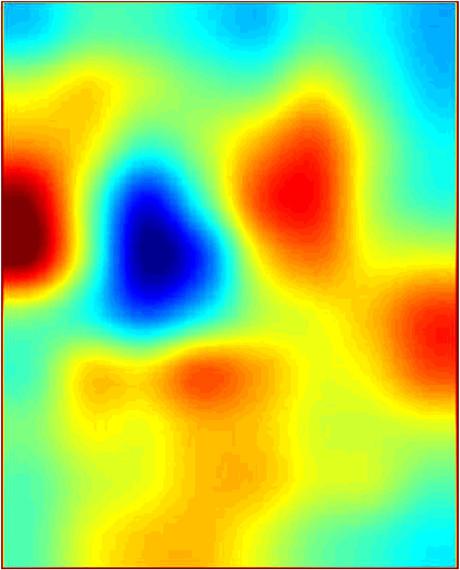}
\end{tabular}&
  \begin{tabular}{cc}
    \includegraphics[height = 0.08\linewidth]{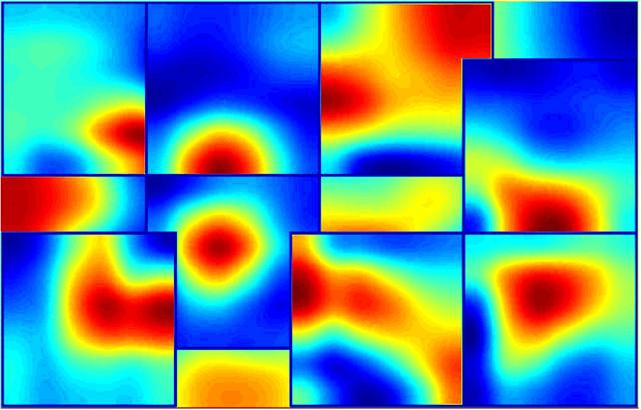}\\
    \includegraphics[height = 0.1\linewidth]{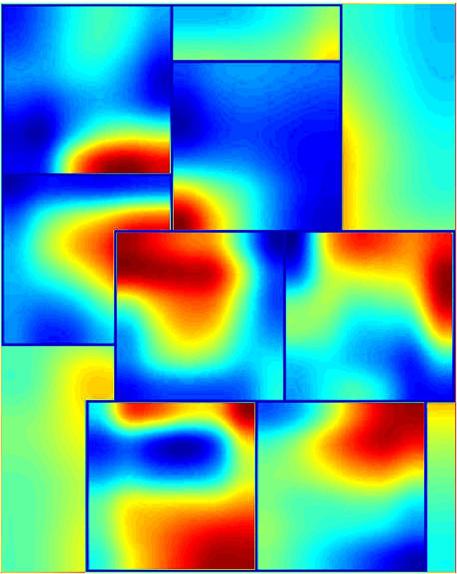}
\end{tabular}&
  \begin{tabular}{cc}
    \includegraphics[height = 0.08\linewidth]{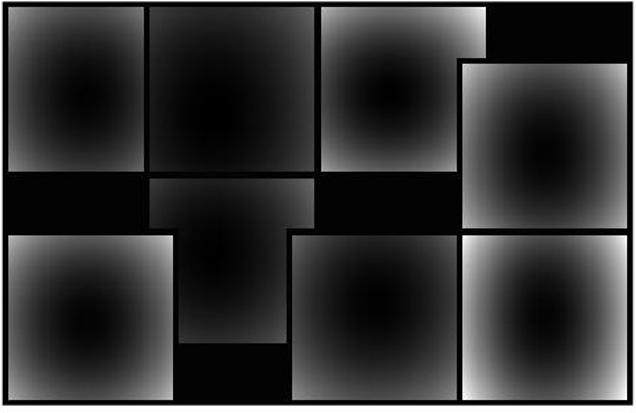}\\
    \includegraphics[height = 0.1\linewidth]{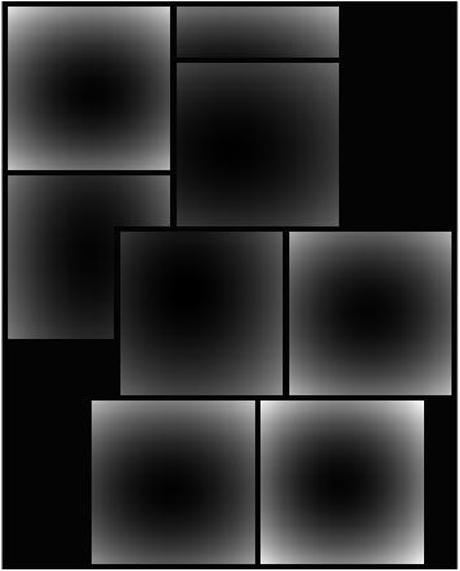}
\end{tabular}\\

\end{tabular}
\caption{Comparison of marginal and component conditioned fixation density maps and corresponding GazeDPM models (column 3 to 7). In the visualizations of gaze filters color close to blue represents negative values, close to red - positive values.}
\label{fig:cat-distribution-components} 
\end{table*}

It is also well-known that different DPM components correspond roughly to different viewpoints on an object.
To analyse this for our model, we computed fixation density maps conditioned on category and associated component of the corresponding DPM models and compared them to fixation density maps only conditioned on the category.
The sample distributions for the ``cat'' class in~\autoref{fig:cat-distribution-components}
show that the component-conditioned fixation density maps and the learnt models differ for the two component model as the component conditioned densities contain view-specific information.
Specifically, the mode of the fixation density map is located in the upper half where the head of animals are usually located.
For different viewpoints the mode location and thus fixation distributions change.
Due to the early integration, our \abbrev~model can exploit this information and we attribute part of its success to the viewpoint-specific fixation density modelling.
This can also be seen by comparing gaze parts of the two component model with gaze parts of the one component model.
Part gaze filters for one component have weights distributed more evenly in order to account for different distributions of different views on object.
On a coarser scale of root filters the opposite dependency holds, as fixation distribution is more stable for coarser scale among different views and thus is more useful for the single cluster model.

\subsection{Performance on Fixation Subsets}

Gaze information available in the POET dataset is collected from five observers.
In many practical applications only a smaller amount of fixation information might be available, such as only from one user or collected for a shorter amount of time.
To study performance of the \abbrev~in these conditions, we run a series of studies on subsets of fixations available for each image in the POET dataset. 

\subsubsection{Influence of Number of Fixations}

We first sampled a random subset of fixations from all available fixations for an image and used these to generate fixation density maps.
\abbrev~was trained and evaluated on these fixation maps and the results are shown in~\autoref{fig-POET-sampling}.
We found that using only 11 fixation -- which is less than half the available amount in POET -- only leads to a reduction of $1\%$ in mAP compared to using all available fixations, which is still an $3\%$ improvement compared to the DPM baseline.
Notably, even only three fixations can already be helpful to yield more than $1\%$ improvement compared to the baseline DPM. For a more complete account of the experiments, please refer to the supplementary material.

\subsubsection{Influence of Order of Fixations}
To investigate the importance of the order of fixations we further sub-sampled fixations but keeping their temporal order.
We then trained and evaluated the performance of \abbrev~with the first $ n \in \{1,3,7,11,15,19,23\} $ fixations and with last $ n \in \{1,3,7,11,15,19,23\} $ fixations.
As shown in~\autoref{fig-POET-sampling}, the last 7 fixations are more informative than the first 7 fixations.
It turns out that the last fixations are more likely to be on the target object due to visual search of the observers.
In particular, using the last 7 fixations -- which is a third of all available fixations -- already results in more than $2\%$ improvement compared to the baseline DPM.

\subsubsection{Influence of User-Specific Fixations}

\begin{figure*}[t]
\centering
\resizebox{0.85\textwidth}{!}{%
\definecolor{mycolor1}{rgb}{0.00000,1.00000,1.00000}%
\definecolor{mycolor2}{rgb}{1.00000,0.00000,1.00000}%
\begin{tikzpicture}

\begin{axis}[%
width=3.220859in,
height=2in,
at={(0in,0in)},
scale only axis,
every outer x axis line/.append style={black},
every x tick label/.append style={font=\color{black}},
xmin=1,
xmax=23,
xtick={ 1,  3,  7, 11, 15, 19, 23},
xlabel={Number of fixations},
every outer y axis line/.append style={black},
every y tick label/.append style={font=\color{black}},
ymin=30,
ymax=35,
ylabel={mAP},
ymajorgrids,
axis x line*=bottom,
axis y line*=left,
legend style={at={(1.03,1)},anchor=north west,legend cell align=left,align=left,draw=black}
]

\addplot [color=mycolor2,solid,line width=2pt]
  table[row sep=crcr]{%
1	34.7\\
23	34.7\\
};
\addlegendentry{All fixations};

\addplot [color=blue,solid,mark=asterisk,mark options={solid},line width=2pt]
  table[row sep=crcr]{%
1	31.1\\
3	31.4\\
7	32.9\\
11	33.9\\
15	33.3\\
19	34\\
23	33.6\\
};
\addlegendentry{Random subsample};

\addplot [color=red,solid,mark=o,mark options={solid},line width=2pt]
  table[row sep=crcr]{%
1	30.7\\
3	32.4\\
7	33.4\\
11	34\\
15	32.6\\
19	33.2\\
23	32.9\\
};
\addlegendentry{Last fixations};

\addplot [color=green,solid,mark=triangle,mark options={solid},line width=2pt]
  table[row sep=crcr]{%
1	31.3\\
3	31.5\\
7	31.7\\
11	31.9\\
15	34.2\\
19	33.9\\
23	33.9\\
};
\addlegendentry{First fixations};

\addplot [color=mycolor1,dashed,line width=2pt]
  table[row sep=crcr]{%
1	31.3\\
23	31.3\\
};
\addlegendentry{BMS v2};

\addplot [color=red,dashed,line width=2pt]
  table[row sep=crcr]{%
1	31.2\\
23	31.2\\
};
\addlegendentry{GBVS};

\addplot [color=black,dash pattern=on 1pt off 3pt on 3pt off 3pt,line width=2pt]
  table[row sep=crcr]{%
1	30.4\\
23	30.4\\
};
\addlegendentry{DPM};

\end{axis}
\end{tikzpicture}%
}
\caption{Performance of \abbrev~in terms of mAP using images of POET dataset and fixation density maps, either estimated using graph-based visual saliency (GBVS) \cite{harel2006graph} or boolean map saliency (BMS) v2 \cite{zhang2013saliency}, or generated from fixations sampled by different criteria from all available fixations for an image.} 
\label{fig-POET-sampling}
\end{figure*}
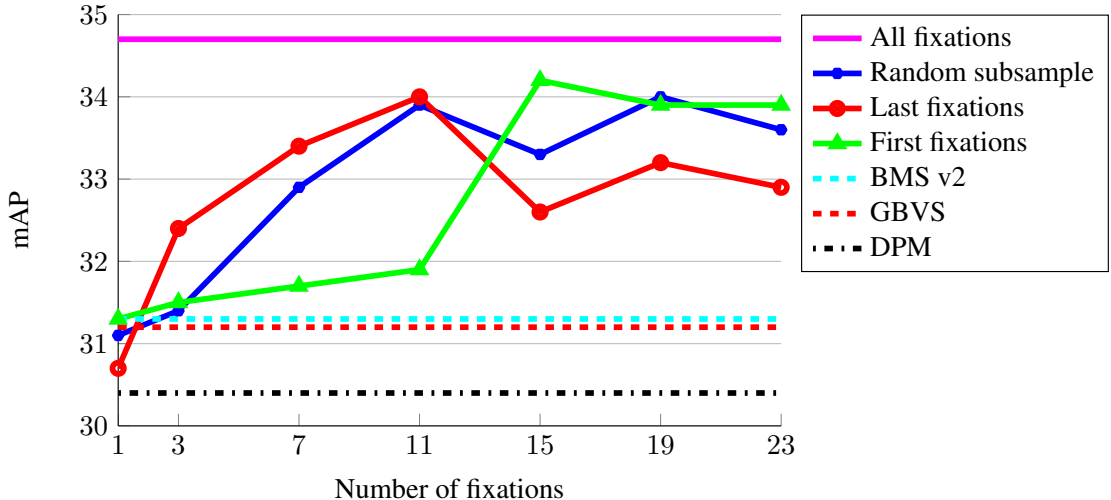

In many practical use cases, only fixations for a small number of users -- most often just a single user -- are available. Consequently, we trained \abbrev~models on fixation maps generated using fixations of a specific user.
Performance results of these models are shown in~\autoref{table-POET-results}.
On average, we got an improvement of 2\% mAP in the single user setting over the baseline DPM.
Note that similar improvement is obtained with fixation maps generated from 5 randomly sampled fixations (see~\autoref{fig-POET-sampling}) which is an average number of fixations in POET dataset for a single observer per image. This suggests that fixations for different observers are roughly equally informative.
Although the average performance of \abbrev~is about equal for different users, for specific classes performance can be quite different (e.g. ``aeroplane'' category performance for user 5).
This suggests biases in individual fixation patterns or search strategies for specific users.
Additional experiments revealed that training on as little as two users can be enough for performance to be only $1\%$ below training on all users.

\subsection{Robustness to Gaze Estimation Error}

All results that we showed so far were obtained using fixation data that is subject to some small amount of noise.
The noise is caused by the inherent and inevitable gaze estimation error in the eye tracker used to record the data.
Data collection in~\cite{yun2013studying} was performed using a high-accuracy remote eye tracker.
However, for many practical applications other sensors become increasingly interesting, such as ordinary monocular RGB cameras that are readily integrated into many mobile phones, laptops, and interactive displays~\cite{zhang14_ubicomp,xucong15cvpr,wood14_etra}.
In these settings, fixation information can be expected to contain substantially more noise due to even lower gaze estimation accuracy of these methods.
We therefore analyzed how robust our \abbrev~model is to simulated noise in the fixation data.

The method proposed in~\cite{sewell2010real} achieved a gaze estimation accuracy of roughly $3 \pm 3 $ degrees of visual angle.
We used this accuracy as a starting point for our investigations of robustness to gaze estimation error.
Note, though, that this accuracy highly depends on the hardware and scenario.
In~\cite{sewell2010real} observers were seated 75cm away from a screen with a size of 28 cm by 18 cm.
We assumes that POET images were scaled proportionally to fit on the screen.
Accordingly, we first translated fixation coordinates for all images into the centimeters of the screen.
We then translated accuracy in degrees to accuracy in centimeters of screen surface, which is $ 75 \tan(3) \pm 75 \tan(3) $.
We used Gaussian noise to approximately model the noise due to gaze estimation error.
Specifically, as distribution parameters of the Gaussian noise we selected $\mu = 0$ and by rule of 3 sigmas we set $ \sigma^2 = \frac{75 \tan(3+3)}{3} $, such that most of resulting Gaussian noise would result in at most 6 degrees of visual angle error.
For comparison, we also considered fixations with added Gaussian noise for different multiples of $ \sigma^2 $.
We generated noise under these assumptions and added it to fixation coordinates, expressed in screen coordinate system. Then we computed fixation density maps from these noisy fixations and trained \abbrev~on them.

Results of this evaluation are shown in~\autoref{table-POET-results}.
Even with a noise level of $\sigma^2$ we still obtain improvement of around 1\% compared to the baseline DPM. 
Such $\sigma^2$ roughly corresponds to an average shift of fixation coordinates of approximately $\pm$ 20 \% pixels of the image height.
With larger values of $\sigma^2$ the improvement vanishes.
For a smaller noise level of  $0.5 \sigma^2$ we get a performance within $0.4\%$ of the measured fixations without noise.
This shows that our \abbrev~method is robust to small noise levels and yields improvement up to medium noise levels.

\subsection{Experiments Using Saliency Maps}

Although the core of our investigation is centered around the use of real fixation data, we finally investigated if our method can also be used in the absence of such data, i.e.\ if 
the fixation density maps are replaced with saliency maps calculated using state-of-the-art saliency methods.%
Specifically, we used 
graph-based visual saliency (GBVS)~\cite{harel2006graph} and boolean map saliency (BMS)~\cite{zhang2013saliency}, which both perform very well by different metrics on the MIT saliency benchmark~\cite{mit-saliency-benchmark}.

As can be seen from~\autoref{fig-POET-sampling} our \abbrev~model achieved an improvement of 0.8\% mAP for GBVS and 1\% mAP using BMS compared to the baseline DPM.
We hypothesize that improvements stem from global features in the saliency map that the local HOG descriptor in the DPM does not have access to.
We also observed that the obtained improvement is roughly consistent with the improvement obtained by one real fixation. %
Although both saliency maps performed comparable in this setting, for some object categories like ``cat'' there was a significant performance difference of up to  10\% (see supplementary material). 

\section{Conclusion}

In this work we have presented an early integration method that improves visual object class detection using human fixation information.
At the core of our approach 
we have proposed the \abbrev~ as an extension to the well-known deformable part model that constitutes a joint formulation over visual and fixation information. 
We have obtained an improvement of 4.3\% of mAP compared to a baseline DPM and around 3.9\% compared to a recent late integration approach.
Further, we have studied a range of cases of practical relevance that are characterized by limited or noisy eye fixation data and observe that our approach is robustness to many such variations which argues for its particability. Besides the quantitative results, we have found that the intraspection gained by visualizing the trained models has led to interesting insights and opens an avenue to further study and understand the interplay between fixation strategies and object cognition.

{\small
\bibliographystyle{ieee}
\bibliography{references}
}

\clearpage

\section{Supplementary Material}

\begin{table*}[h]
\begin{center}
\begin{tabular}{lllll}
\textbf{Clusters} & \textbf{DPM} & \textbf{Gaze} & \textbf{Noise} & \textbf{Zero} \\ \hline
2 & 30.2 & 34.7 & 30.8 & 29.8 \\ \hline
3 & 29.9 & 34.5 & 29.6 & 29.7 \\ \hline
4 & 27.6 & 31.5 & 28.0 & 28.4 \\ 
\end{tabular}
\end{center}
\caption{Comparison of performance of different modifications of DPM with different number of clusters. 'Gaze' corresponds to Gaze DPM used with fixation maps generated from real fixations,  'Noise' corresponds to Gaze DPM used with fixation map where every value is set to zero, 'DPM' is unchanged implementation of dpm library \cite{felzenszwalb2010object}, 'Noise' corresponds to performance of Gaze DPM with fixation maps filled with uniform noise.}
\label{table-cluster-number}
\end{table*}

\begin{table*}[h]
\begin{center}
\begin{tabular}{lllllllll}
\textbf{Class} & \textbf{DPM} & \textbf{\begin{tabular}[c]{@{}l@{}}100 ms \\ (1fx)\end{tabular}} & \textbf{\begin{tabular}[c]{@{}l@{}}200 ms\\  (3fx)\end{tabular}} & \textbf{\begin{tabular}[c]{@{}l@{}}300 ms \\ (9fx)\end{tabular}} & \textbf{\begin{tabular}[c]{@{}l@{}}400 ms \\ (13fx)\end{tabular}} & \textbf{\begin{tabular}[c]{@{}l@{}}600 ms \\ (18fx)\end{tabular}} & \textbf{\begin{tabular}[c]{@{}l@{}}800 ms \\ (22fx)\end{tabular}} & \textbf{All} \\ \hline
cat & 23.9 & 28.8 & 32.1 & 37.7 & 37.5 & 37.0 & 37.5 & 40.2 \\ \hline
cow & 22.6 & 20.3 & 23.3 & 21.9 & 20.6 & 23.7 & 24.5 & 24.9 \\ \hline
dog & 14.7 & 18.1 & 15.5 & 21.1 & 23.8 & 25.3 & 27.7 & 28.2 \\ \hline
horse & 43.9 & 43.5 & 44.7 & 45.3 & 43.7 & 45.3 & 45.6 & 46.0 \\ \hline
aeroplane & 41.8 & 45.6 & 44.6 & 43.0 & 41.8 & 42.4 & 43.8 & 40.6 \\ \hline
bicycle & 53.5 & 52.2 & 53.1 & 53.6 & 53.7 & 55.8 & 56.6 & 53.5 \\ \hline
boat & 8.4 & 7.3 & 8.2 & 10.2 & 8.5 & 9.9 & 10.2 & 9.3 \\ \hline
diningtable & 19.8 & 18.8 & 19.1 & 24.4 & 27.4 & 22.7 & 26.9 & 30.0 \\ \hline
motorbike & 48.5 & 46.9 & 45.2 & 45.1 & 44.8 & 44.0 & 45.6 & 45.9 \\ \hline
sofa & 26.7 & 27.3 & 31.5 & 26.6 & 30.4 & 28.5 & 32.3 & 28.5 \\ \hline
\textbf{Average} & \textbf{30.4} & \textbf{30.9} & \textbf{31.7} & \textbf{32.9} & \textbf{33.2} & \textbf{33.5} & \textbf{35.1} & \textbf{34.7} \\ 
\end{tabular}
\end{center}
\caption{Performance of gaze enabled dpm modification in terms of mAP on images of POET dataset using different number of fixations sampled until a certain viewing time. For each column, the corresponding average number of fixations for viewing time is specified.} 
\label{table-POET-until-time}
\end{table*}
\begin{table*}[h]
\begin{center}
\begin{tabular}{lllllllll}
\textbf{Class} & \textbf{All} & \textbf{\begin{tabular}[c]{@{}l@{}}100 ms \\ (24fx)\end{tabular}} & \textbf{\begin{tabular}[c]{@{}l@{}}200 ms \\ (22fx)\end{tabular}} & \textbf{\begin{tabular}[c]{@{}l@{}}300 ms \\ (17fx)\end{tabular}} & \textbf{\begin{tabular}[c]{@{}l@{}}400 ms \\ (12fx)\end{tabular}} & \textbf{\begin{tabular}[c]{@{}l@{}}600 ms \\ (8fx)\end{tabular}} & \textbf{\begin{tabular}[c]{@{}l@{}}800 ms \\ (3fx)\end{tabular}} & \textbf{DPM} \\ \hline
cat & 40.2 & 37.6 & 37.4 & 38.6 & 38.6 & 37.0 & 28.5 & 23.9 \\ \hline
cow & 24.9 & 23.6 & 22.3 & 26.0 & 22.2 & 24.5 & 23.2 & 22.6 \\ \hline
dog & 28.2 & 28.9 & 26.0 & 24.1 & 22.3 & 15.5 & 20.9 & 14.7 \\ \hline
horse & 46.0 & 45.7 & 45.2 & 44.4 & 45.6 & 43.8 & 42.9 & 43.9 \\ \hline
aeroplane & 40.6 & 41.7 & 42.4 & 41.9 & 40.2 & 43.4 & 46.7 & 41.8 \\ \hline
bicycle & 53.5 & 55.5 & 55.8 & 53.4 & 53.8 & 53.7 & 53.8 & 53.5 \\ \hline
boat & 9.3 & 10.4 & 9.5 & 10.7 & 9.7 & 9.7 & 7.5 & 8.4 \\ \hline
diningtable & 30.0 & 26.6 & 22.6 & 30.0 & 27.7 & 26.3 & 24.9 & 19.8 \\ \hline
motorbike & 45.9 & 46.2 & 42.7 & 44.6 & 43.6 & 47.5 & 46.8 & 48.5 \\ \hline
sofa & 28.5 & 31.5 & 30.9 & 29.2 & 28.8 & 24.4 & 27.8 & 26.7 \\ \hline
\textbf{Average} & \textbf{34.7} & \textbf{34.8} & \textbf{33.5} & \textbf{34.3} & \textbf{33.2} & \textbf{32.6} & \textbf{32.3} & \textbf{30.4} \\ 
\end{tabular}
\end{center}
\caption{Performance of gaze enabled dpm modification in terms of mAP on images of POET dataset using different number of fixations sampled after a certain viewing time. For each column, the corresponding average number of fixations for viewing time is specified.} 
\label{table-POET-after-time}
\end{table*}

\begin{table*}[h]
\begin{center}
\begin{tabular}{lllllllllll}
\textbf{Class} & \textbf{DPM} & \textbf{1 fix-s} & \textbf{2 fix-s} & \textbf{3 fix-s} & \textbf{7 fix-s} & \textbf{11 fix-s} & \textbf{15 fix-s} & \textbf{19 fix-s} & \textbf{23 fix-s} & \textbf{All} \\ \hline
\multicolumn{1}{l}{cat} & 23.9 & 32.9 & 35.3 & 36.1 & 35.6 & 37.5 & 36.4 & 37.3 & 38.2 & 40.2 \\ \hline
\multicolumn{1}{l}{cow} & 22.6 & 19.7 & 21.4 & 18.9 & 22.4 & 24.0 & 21.9 & 22.0 & 23.7 & 24.9 \\ \hline
\multicolumn{1}{l}{dog} & 14.7 & 19.8 & 22.7 & 24.5 & 21.9 & 24.9 & 26.2 & 26.1 & 25.1 & 28.2 \\ \hline
\multicolumn{1}{l}{horse} & 43.9 & 44.5 & 43.6 & 43.9 & 43.4 & 43.8 & 42.3 & 43.9 & 44.1 & 46.0 \\ \hline
\multicolumn{1}{l}{aeroplane} & 41.8 & 42.7 & 42.8 & 39.5 & 42.3 & 40.2 & 39.3 & 40.9 & 40.6 & 40.6 \\ \hline
\multicolumn{1}{l}{bicycle} & 53.5 & 51.6 & 54.0 & 50.8 & 53.1 & 54.5 & 53.6 & 55.7 & 55.3 & 53.5 \\ \hline
\multicolumn{1}{l}{boat} & 8.4 & 6.0 & 7.4 & 6.2 & 9.8 & 9.6 & 9.4 & 10.0 & 9.8 & 9.3 \\ \hline
\multicolumn{1}{l}{diningtable} & 19.8 & 21.3 & 22.7 & 22.8 & 26.2 & 29.3 & 27.4 & 27.8 & 24.6 & 30.0 \\ \hline
\multicolumn{1}{l}{motorbike} & 48.5 & 47.2 & 42.2 & 45.4 & 46.4 & 46.6 & 45.2 & 45.5 & 46.2 & 45.9 \\ \hline
\multicolumn{1}{l}{sofa} & 26.7 & 24.8 & 24.9 & 26.1 & 28.2 & 28.3 & 30.9 & 31.1 & 28.7 & 28.5 \\ \hline
\multicolumn{1}{l}{\textbf{Average}} & \textbf{30.4} & \textbf{31.1} & \textbf{31.7} & \textbf{31.4} & \textbf{32.9} & \textbf{33.9} & \textbf{33.3} & \textbf{34.0} & \textbf{33.6} & \textbf{34.7} \\ 
\end{tabular}
\end{center}
\caption{Performance of gaze enabled dpm modification in terms of mAP on images of POET dataset using different number of fixations sampled randomly from all available fixations.} 
\label{table-POET-random-sampling}
\end{table*}

\begin{table*}[h]
\begin{center}
\begin{tabular}{llllllllll}
\textbf{Class} & \textbf{DPM} & \textbf{1 fx} & \textbf{3 fx} & \textbf{7 fx} & \textbf{11 fx} & \textbf{15 fx} & \textbf{19 fx} & \textbf{23 fx} & \textbf{All} \\ \hline
cat & 23.9 & 29.4 & 29.8 & 35.8 & 33.7 & 38.4 & 38.3 & 36.4 & 40.2 \\ \hline
cow & 22.6 & 24.0 & 19.2 & 22.0 & 19.8 & 21.9 & 24.4 & 23.2 & 24.9 \\ \hline
dog & 14.7 & 18.4 & 18.3 & 18.1 & 20.2 & 26.0 & 24.1 & 25.4 & 28.2 \\ \hline
horse & 43.9 & 43.1 & 44.5 & 45.3 & 46.1 & 47.0 & 45.4 & 44.0 & 46.0 \\ \hline
aeroplane & 41.8 & 46.5 & 42.7 & 44.7 & 43.9 & 43.0 & 42.8 & 43.1 & 40.6 \\ \hline
bicycle & 53.5 & 52.5 & 54.7 & 54.7 & 53.2 & 54.2 & 54.7 & 55.7 & 53.5 \\ \hline
boat & 8.4 & 8.1 & 9.3 & 9.0 & 11.1 & 9.6 & 10.7 & 10.9 & 9.3 \\ \hline
diningtable & 19.8 & 20.3 & 22.8 & 19.7 & 22.9 & 23.2 & 23.4 & 21.9 & 30.0 \\ \hline
motorbike & 48.5 & 46.6 & 47.6 & 43.6 & 44.1 & 46.6 & 45.8 & 47.1 & 45.9 \\ \hline
sofa & 26.7 & 24.0 & 26.3 & 23.6 & 24.1 & 32.1 & 29.3 & 31.5 & 28.5 \\ \hline
\textbf{Average} & \textbf{30.4} & \textbf{31.3} & \textbf{31.5} & \textbf{31.7} & \textbf{31.9} & \textbf{34.2} & \textbf{33.9} & \textbf{33.9} & \textbf{34.7} \\ 
\end{tabular}
\end{center}
\caption{Performance of gaze enabled dpm modification in terms of mAP on images of POET dataset using different number of first fixations.} 
\label{table-POET-first-number}
\end{table*}
\begin{table*}[h]
\begin{center}
\begin{tabular}{llllllllll}
\textbf{Class} & \textbf{DPM} & \textbf{1 fx} & \textbf{3 fx} & \textbf{7 fx} & \textbf{11 fx} & \textbf{15 fx} & \textbf{19 fx} & \textbf{23 fx} & \textbf{All} \\ \hline
cat & 23.9 & 33.8 & 35.1 & 34.7 & 39.7 & 36.6 & 35.5 & 38.7 & 40.2 \\ \hline
cow & 22.6 & 20.4 & 19.2 & 21.9 & 22.0 & 21.8 & 19.8 & 22.4 & 24.9 \\ \hline
dog & 14.7 & 14.8 & 22.4 & 24.9 & 23.1 & 23.6 & 28.6 & 26.7 & 28.2 \\ \hline
horse & 43.9 & 43.6 & 43.2 & 43.3 & 44.2 & 44.8 & 46.4 & 45.7 & 46.0 \\ \hline
aeroplane & 41.8 & 43.9 & 45.3 & 47.0 & 41.9 & 41.2 & 42.4 & 39.5 & 40.6 \\ \hline
bicycle & 53.5 & 52.8 & 53.2 & 51.7 & 53.8 & 54.0 & 54.6 & 54.6 & 53.5 \\ \hline
boat & 8.4 & 8.1 & 8.1 & 9.5 & 11.4 & 9.1 & 10.5 & 11.8 & 9.3 \\ \hline
diningtable & 19.8 & 17.3 & 25.9 & 26.9 & 28.7 & 19.8 & 21.9 & 22.7 & 30.0 \\ \hline
motorbike & 48.5 & 47.1 & 45.8 & 45.9 & 47.4 & 44.1 & 44.8 & 43.5 & 45.9 \\ \hline
sofa & 26.7 & 25.5 & 26.0 & 28.1 & 27.6 & 30.8 & 27.2 & 23.6 & 28.5 \\ \hline
\textbf{Average} & \textbf{30.4} & \textbf{30.7} & \textbf{32.4} & \textbf{33.4} & \textbf{34.0} & \textbf{32.6} & \textbf{33.2} & \textbf{32.9} & \textbf{34.7} \\ 
\end{tabular}
\end{center}
\caption{Performance of gaze enabled dpm modification in terms of mAP on images of POET dataset using different number of last fixations.} 
\label{table-POET-last-number}
\end{table*}

\begin{table*}[t]
\begin{center}
\begin{tabular}{llll}
\textbf{Class} & \textbf{DPM} & \textbf{GBVS} & \textbf{BMSV2} \\ \hline
cat & 23.9 & 30.2 & 30.2 \\ \hline
cow & 22.6 & 21.4 & 23.8 \\ \hline
dog & 14.7 & 24.5 & 17.5 \\ \hline
horse & 43.9 & 40.0 & 42.6 \\ \hline
aeroplane & 41.8 & 40.9 & 40.5 \\ \hline
bicycle & 53.5 & 53.3 & 53.9 \\ \hline
boat & 8.4 & 9.5 & 7.4 \\ \hline
diningtable & 19.8 & 20.4 & 22.3 \\ \hline
motorbike & 48.5 & 44.4 & 48.9 \\ \hline
sofa & 26.7 & 27.7 & 25.7 \\ \hline
\textbf{Average} & \textbf{30.4} & \textbf{31.2} & \textbf{31.3} \\ 
\end{tabular}
\end{center}
\caption{Performance of Gaze DPM on images of POET dataset with generated fixation maps using BMS V2, using maps from BMS corresponding to salient object detection, maps generated from POET fixations and unmodified dpm library. For all modifications 2 aspect ratio clusters are used. 'DPM' is unchanged implementation of dpm library. } 
\end{table*}

\subsection{Multiple fixation maps}

In this sections, experiments are described where instead of single fixation map generated from fixations multiple fixation maps are used, generated from fixations separated by certain criterion, like fixation from certain viewing time.

We normalize viewing time for every user separately, to account for possible differences in user reaction time. Specifically, for all fixations for specific user, we compute average over all viewing times (as viewing time we use time of the end of last fixation) for all images, normalize all fixation times using this value. We do not use viewing time for 3 images in each class that were viewed first, to avoid outliers.

\subsubsection{Fixation length and viewing time as separation criterion}

Soft binning was used to separate fixations for different saliency maps channels. Fixation length is normalized by viewing time to account for differences in reaction time, and results are shown in Table \ref{table-fx-hard-binning-fixation-duration-time-norm}. K-means was used to determine clusters based on fixation duration and viewing time when fixation was made.

\begin{table}[h]
\begin{center}
\begin{tabular}{llllll}
\textbf{Class} & \textbf{DPM} & \textbf{2 cl.} & \textbf{3 cl.} & \textbf{4 cl.} & \textbf{GazeDPM} \\ \hline
cat & 23.9 & 37.4 & 36.7 & 36.3 & 40.2 \\ \hline
cow & 22.6 & 22.0 & 21.2 & 19.9 & 24.9 \\ \hline
dog & 14.7 & 23.6 & 26.2 & 23.5 & 28.2 \\ \hline
horse & 43.9 & 41.9 & 44.5 & 41.4 & 46.0 \\ \hline
aeroplane & 41.8 & 36.9 & 40.4 & 32.4 & 40.6 \\ \hline
bicycle & 53.5 & 52.8 & 53.1 & 52.3 & 53.5 \\ \hline
boat & 8.4 & 11.1 & 10.2 & 8.6 & 9.3 \\ \hline
diningtable & 19.8 & 17.5 & 17.5 & 12.8 & 30.0 \\ \hline
motorbike & 48.5 & 41.9 & 41.0 & 40.6 & 45.9 \\ \hline
sofa & 26.7 & 26.4 & 29.4 & 23.1 & 28.5 \\ \hline
\textbf{Average} & \textbf{30.4} & \textbf{31.1} & \textbf{32.0} & \textbf{29.1} & \textbf{34.7} \\ 
\end{tabular}
\caption{Performance of gaze enabled dpm modification in terms of mAP on images of POET dataset using different number of fixations saliency map features, compared to baseline performance (no gaze information, DPM coumn) and when single fixation map feature is used (GazeDPM column). K-means was used to determine clusters based on fixation duration and viewing time when fixation was made. Fixation duration for specific participant of POET data collection is normalized by average viewing time. } 
\label{table-fx-hard-binning-fixation-duration-time-norm}
\end{center}
\end{table}

\subsubsection{Fixation length as separation criterion}

Soft binning was used to separate fixations for different saliency maps channels. Fixation length is normalized by viewing time to account for differences in reaction time, and results are shown in Table \ref{table-fx-hard-binning-fixation-time-norm}.  K-means was used to determine clusters based on fixation duration.

\begin{table}[h]
\begin{center}
\begin{tabular}{lllll}
\textbf{Class} & \textbf{DPM} & \textbf{2 cl.} & \textbf{3 cl.} & \textbf{GazeDPM} \\ \hline
cat & 23.9 & 37.4 & 36.7 & 40.2 \\ \hline
cow & 22.6 & 22.6 & 21.6 & 24.9 \\ \hline
dog & 14.7 & 26.1 & 24.5 & 28.2 \\ \hline
horse & 43.9 & 42.5 & 42.8 & 46.0 \\ \hline
aeroplane & 41.8 & 43.4 & 37.6 & 40.6 \\ \hline
bicycle & 53.5 & 54.5 & 53.2 & 53.5 \\ \hline
boat & 8.4 & 9.2 & 8.3 & 9.3 \\ \hline
diningtable & 19.8 & 25.2 & 22.1 & 30.0 \\ \hline
motorbike & 48.5 & 46.1 & 42.4 & 45.9 \\ \hline
sofa & 26.7 & 29.3 & 26.9 & 28.5 \\ \hline
\textbf{Average} & \textbf{30.4} & \textbf{33.6} & \textbf{31.6} & \textbf{34.7} \\ 
\end{tabular}
\caption{Performance of gaze enabled dpm modification in terms of mAP on images of POET dataset using different number of fixations saliency map features, compared to baseline performance (no gaze information, DPM coumn) and when single fixation map feature is used (GazeDPM column). K-means was used to determine clusters based on fixation duration. Fixation duration for specific participant of POET data collection is normalized by average viewing time. } 
\label{table-fx-hard-binning-fixation-time-norm}
\end{center}
\end{table}

As the number of fixations per image is small, soft binning based on similarity was used. Centroids from k-means algorithm are used as centers of bins for certain gaze feature. To determine contriution of certain fixation to a bin with centroid $ c \in C $, where $  C \subset \mathbb{R}^n $ is a set of $n \in \mathbb{N}$ centroids, the following formula is used
\begin{align}
a(d, c) &= \dfrac{s(d, c)}{\sum\limits_{c' \in C} s(d, c')} \\
s(d, c) &= \mathcal{N}(d,\sigma)(c) = \frac{1}{{\sigma \sqrt {2\pi } }}e^{{{ - \left( {x - \mu } \right)^2 } \mathord{\left/ {\vphantom {{ - \left( {d - c } \right)^2 } {2\sigma ^2 }}} \right. \kern-\nulldelimiterspace} {2\sigma ^2 }}} \\
\sigma &= 0.025
\end{align}
where $ d $ is a fixation length, $ c $ - centroid that corresponds to a certain saliency map feature, $  a(d, c) $ - contribution of fixation with length $ d $ to gaze feature with centroid $ c $.

\end{document}